\documentclass[10pt,twocolumn,letterpaper]{article}

\usepackage{iccv}
\usepackage[title,titletoc]{appendix}
\usepackage{times}
\usepackage{epsfig}
\usepackage{graphicx}
\usepackage{booktabs}
\usepackage{amssymb}
\usepackage{algorithm}
\usepackage{algorithmic}
\usepackage{amsmath,bm}
\usepackage[table]{xcolor}
\usepackage{arydshln}  

\usepackage[pagebackref=false,breaklinks=false,letterpaper=true,colorlinks,bookmarks=false,citecolor=green, linkcolor=blue]{hyperref}

\iccvfinalcopy 

\def\iccvPaperID{****} 

\ificcvfinal\fi

\begin{document}

\title{Small Object Detection via Coarse-to-fine Proposal Generation \\ and Imitation Learning}

\author{
	Xiang Yuan\quad Gong Cheng \thanks{Corresponding author: gcheng@nwpu.edu.cn} \quad Kebing Yan\quad Qinghua Zeng\quad Junwei Han \\
	School of Automation, Northwestern Polytechnical University, Xi’an, China \\
	{\tt\small $\lbrace$shaunyuan, kebingyan, zengqinghua$\rbrace$@mail.nwpu.edu.cn,}
	{\tt\small  $\lbrace$gcheng, jhan$\rbrace$@nwpu.edu.cn}
}

\maketitle
\ificcvfinal\thispagestyle{empty}\fi

\begin{abstract}
	The past few years have witnessed the immense success of object detection, while current excellent detectors struggle on tackling size-limited instances. 
	Concretely, the well-known challenge of low overlaps between the priors and object regions leads to a constrained sample pool for optimization, and the paucity of discriminative information further aggravates the recognition.
	To alleviate the aforementioned issues, we propose CFINet, a two-stage framework tailored for small object detection based on the Coarse-to-fine pipeline and Feature Imitation learning.
	Firstly, we introduce Coarse-to-fine RPN (CRPN) to ensure sufficient and high-quality proposals for small objects through the dynamic anchor selection strategy and cascade regression. 
	Then, we equip the conventional detection head with a Feature Imitation (FI) branch to facilitate the region representations of size-limited instances that perplex the model in an imitation manner.  
	Moreover, an auxiliary imitation loss following supervised contrastive learning paradigm is devised to optimize this branch.
	When integrated with Faster RCNN, CFINet achieves state-of-the-art performance on the large-scale small object detection benchmarks, SODA-D and SODA-A, underscoring its superiority over baseline detector and other mainstream detection approaches.
\end{abstract}

\vspace{-1em}
\section{Introduction}

Small object detection (SOD) \footnote[1]{This paper focuses on the detection of "\textbf{pure}" small objects, where the scales of all the objects are distributed within a relatively tight range \cite{tinyperson, aitod, soda}, which is distinct from the known small objects in multi-scale object detection \cite{coco}.} aims to classify and localize the instances with limited regions, which plays an important role in a wide range of scenarios, such as pedestrian detection, autonomous driving, and intelligent surveillance understanding, to name a few \cite{querydet, bffb, pergan, lpr, mimic, mtgan, bam}. 
Compared to the generic object detection which has been extensively studied, SOD task receives relatively little attention and good solutions are still scarce so far. 
Moreover, generic detectors \cite{fasterrcnn, sfrnet, fpn, fcos, aopg, detr, retinanet, rifd} usually struggle on handling small objects due to two inherent challenges: the \textbf{insufficient and low-quality samples for training} and the \textbf{uncertain prediction of RoIs} (Region of Interests).

\begin{figure}[t]
	\centering
	\includegraphics[scale=0.425]{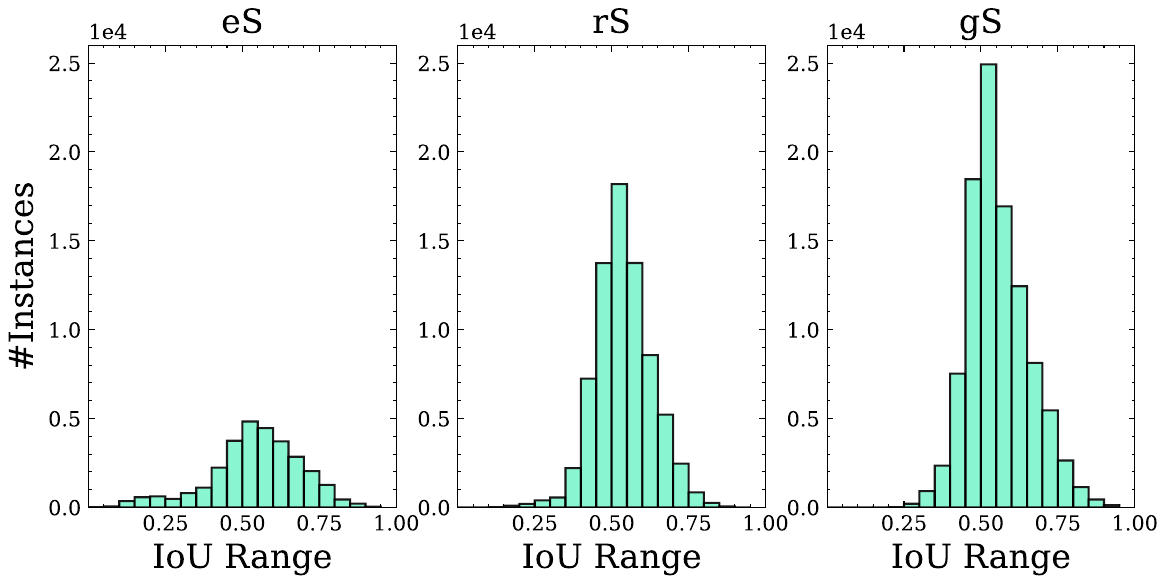} 
	\caption{Distribution of maximum IoU of anchors matched to each ground-truth instance in SODA-D \cite{soda} \texttt{train-set}, where \textit{extremely Small} (\textit{eS}), \textit{relatively Small} (\textit{rS}), and \textit{generally Small} (\textit{gS}) correspond to three area subsets in SODA \cite{soda} with the ranges $\left( 0,144 \right]$,  $\left( 144,400 \right]$ and  $\left( 400,1024 \right]$. The smaller the objects are, the lower IoU the matched anchors have, hence the commonly used positive IoU threshold (0.7) is too rigorous for small objects.}
	\label{fig:Fig1}
	\vspace{-1.5em}
\end{figure}

First, current prevailing detectors exploit either overlap-based \cite{fasterrcnn, retinanet} or distance-based \cite{fcos} strategies to select the positive priors of objects for training. 
However, small instances usually occupy an extremely limited area, therefore the region overlaps between densely arranged anchors and ground truth boxes are significantly small and far from the commonly used positive IoU (Intersection-over-Union) threshold, as in Figure \ref{fig:Fig1}. 
In other words, the existing \textit{positive sample} criterion is overly stringent when applied to small/tiny objects, resulting in a restricted number of samples available for optimization.
An intuitive approach involves reducing the threshold for defining a positive sample \cite{s3fd}. 
However, while this can lead to an increase in the number of positive samples, it often at the expense of overall sample quality, in which low-quality samples disrupt optimization and incur a trivial regression solution. 
Worse still, this is actually contradictory to the purpose of proposal network, \textit{i.e.}, guaranteeing the recall and ease the burden of subsequent work. 

To sum up, current prevailing priors-to-proposals paradigms that heavily depend on the overlap or distance metric have inherent limitations in detecting small objects, and nowadays devised assignment or sampling schemes contribute minimally to this problem \cite{soda, rfla}. 
Since the proposals play such a crucial role in two-stage detectors, so how about the improved Region Proposal Network (RPN) variants \cite{refine, cascaderpn, garpn} meet small object detection?
Following this line, we take Cascade RPN \cite{cascaderpn}, one of the most superior proposal network for generic object detection, to perform preliminary experiments and the results are shown in Table \ref{tab:Tab1}. 
While the auxiliary regression phase provides refined priors with better initialization for subsequent regression, the final results remain somewhat unsatisfactory. 
Specifically, the improvement mainly comes from the larger objects and the $AP_{eS}$ as well as $AP_{rS}$  actually decrease significantly instead, indicating that the region-based sampling strategy is inclined to large instances which further dominate the proposal network.
Meanwhile, enlarging the sample region contributes little (even negative) to this condition (see the bottom row in Table \ref{tab:Tab1}).
Therefore, the coarse-to-fine pipeline has the potential to surmount the barrier of conventional prior-to-proposal paradigm, but the crux lies in dedicating sufficient attention to small instances.

\begin{table}[t]
	\centering
	\small
	\resizebox{\columnwidth}{!}{
		\begin{tabular}{|c|ccc|cccc|}
			\hline
			\textit{\textit{ctr}/\textit{ign} ratio} & $AP$  & $AP_{50}$  & $AP_{75}$  & $AP_{eS}$  & $AP_{rS}$  & $AP_{gS}$ & $AP_{N}$\\
			\hline
			\hline
			Baseline  & 28.9  & \textbf{59.4}  & 24.1  & \textbf{13.8}  & 25.7  & 34.5  & 43.0 \\\
			0.2/0.5  & 29.1	 & 56.5	 & 25.9	 & 12.5	 & 25.5	 & 35.4	 & 44.7 \\
			0.5/0.8  & \textbf{29.5} & 57.8	 & \textbf{26.0}	 & 13.5	 & \textbf{26.2}	 & \textbf{35.8}	 & \textbf{45.0} \\
			0.8/1.0  & 27.5  & 54.1  & 24.3  & 11.2  & 23.8  & 33.7  & 42.5 \\
			\hline
		\end{tabular}%
		\label{tab:Tab1}%
	}
	\vspace{0.5em}
	\caption{The performances of Cascade RPN \cite{cascaderpn} compared to the baseline (a vanilla Faster RCNN \cite{fasterrcnn}). The \textit{ctr}/\textit{ign} ratio denotes the sampling region in first regression stage of Cascade RPN. The results are tested on the SODA-D \cite{soda} \texttt{test-set} and with a ResNet-50 \cite{resnet} as the backbone.}
	\vspace{-1.5em}
\end{table}

Second, small objects usually lack discriminative information and distorted structures, leading the inclination of model to give ambiguous even incorrect predictions \cite{finding, efpn}.
Meanwhile, there are a certain amount of large instances embodying clear visual cues and better discrimination. 
Building upon this observation, several works proposed to bridge the representation gap between small objects and large ones, and most of them \cite{mtgan, pergan, bffb, finding} rely on Generative Adversarial Network (GAN) \cite{gan} or similarity learning \cite{lpr, mimic} to super-resolve/restore the features of size-limited instances under the guidance of large ones that are deemed to be visually authentic. 
However, these approaches overlook the fact: \textbf{high quality} $\neq$ \textbf{large size} meanwhile \textbf{small size} $\neq$ \textbf{low quality}. 
In other words, the criterion for humans and the model to decide whether a sample is competent to be a good example is distinct. 
For the latter, it is dynamic and should be adjusted according to the current optimization of the detector. 
Moreover, efforts in this line have to resort to sophisticated training strategies or additional models, which is time-consuming and break the conventional end-to-end paradigm.

Putting the above parts together, we propose a two-stage small object detector CFINet based on the coarse-to-fine pipeline and feature imitation learning. 
Concretely, enlightened by the multi-stage proposal generation scheme in Cascade RPN, we devise Coarse-to-fine RPN (CRPN). 
It firstly employs an dynamic anchor selection strategy to mine potential priors to conduct coarse regression, and henceforth, these refined anchors will be classified and regressed by the region proposal network. 
In addition, we extend the conventional classification-and-regression setting with an auxiliary Feature Imitation (FI) branch, which can leverage the regional features of high-quality instances to guide the learning of those objects with uncertain/mistaken predictions, and a loss function based on the Supervised Contrastive Learning (SCL) \cite{scl} is designed to optimize the whole process. 
The main contributions of this paper are summarized as follows:
\begin{itemize}
	\item A coarse-to-fine proposal generation pipeline named CRPN was built to perform anchor-to-proposal procedure, where an area-based anchor mining strategy and cascade regression empower the high-quality proposals for small instances.
	\item An auxiliary Feature Imitation (FI) branch was introduced to enrich the representations of low-quality instances perplexing the model under the supervision of high-quality instances, and this novel branch is optimized by a tailored loss function based on SCL.
	\item The experiment results on the SODA-D and SODA-A datasets exhibit the superiority of our CFINet to detect these instances with extremely limited sizes.
\end{itemize}

\begin{figure*}[t]
	\centering
	\includegraphics[scale=1]{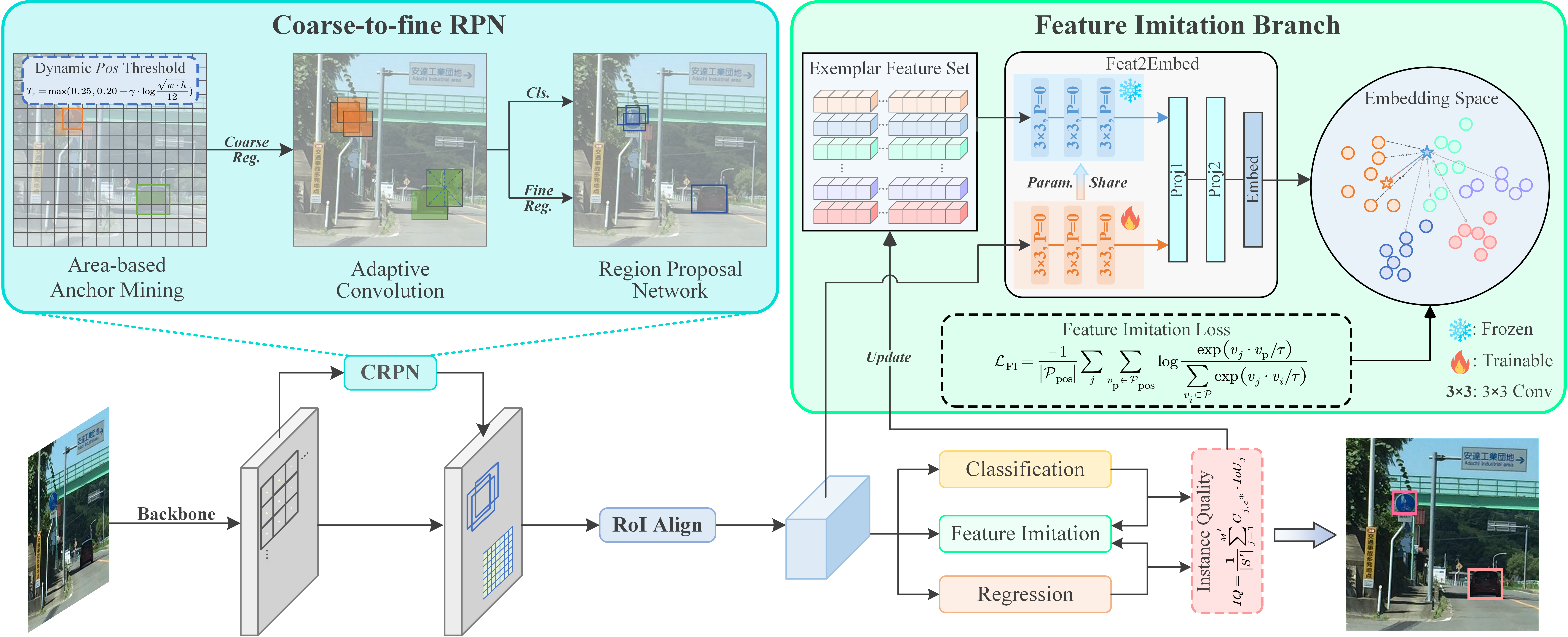} 
	\caption{The overall architecture of CFINet. 
		In Coarse-to-fine RPN (CRPN), the area-based Anchor Mining strategy ensures sufficient candidates for instances of various sizes (small: orange boxes, large: green boxes) based on the dynamic \textit{pos} threshold, which will be then used to obtain the coarse proposals. 
		After that, the alignment between coarse proposals and corresponding features is enabled by the Adaptive Convolution before feeding into the RPN to produce high-quality proposals (blue boxes). 
		The Feature Imitation (FI) branch is devised to facilitate the representations of small instances, in which the RoI features of uncertain/mistaken predictions will be pulled to their counterparts of exemplar feature set in embedding space (throughout the \textit{Feat2Embed} module), while pushed apart from the exemplar features of other categories and background. 
		And the exemplar features are collected based on the model predictions using the proposed quality indicator, \textit{i.e.}, Instance Quality ($IQ$).
		We tailor a Feature Imitation loss function $\mathcal{L}_{\mathrm{FI}}$ to optimize this auxiliary branch. Note that we only exhibit single-level Feature Pyramid Network (FPN) \cite{fpn} feature for clear illustration.
	}
	\label{fig:Fig2}
\end{figure*}

\section{Related Works}
\noindent \textbf{Anchor Refinement and Region Proposals.}
Two-stage anchor-based approaches heavily depend on the high-quality proposals \cite{garpn, cascaderpn}. 
Towards this goal, RPN was first introduced in Faster RCNN \cite{fasterrcnn} to produce proposals in a fully convolutional network, and this simple yet effective design facilitates end-to-end model optimization.
Following RPN, \cite{refine} proposed to iteratively regress the predefined anchors. 
GA-RPN \cite{garpn} discards the uniform anchoring strategy and formulates the anchor generation in two steps: first determining the locations which may contain objects and on which the anchor scales are predicted then. 
By installing a multi-stage anchor-to-proposal strategy and alleviating the misalignment between the refined anchors and image features, Cascade RPN \cite{cascaderpn} enables high-quality proposal generation. 
Unfortunately, current proposal-oriented frameworks fail to produce high-quality proposals for instances with limited regions, and the root cause lies in the notoriously low overlaps between the objects and priors \cite{rfla}. 
Different from the above methods, our coarse-to-fine proposal pipeline could exploit the potential of multi-stage refinement paradigm, thereby guaranteeing both the \textbf{quantity} and \textbf{quality} of proposals for instances with extremely limited sizes.

\noindent \textbf{Feature Imitation for Small Object Detection.}
One of the major challenges to detect small objects is the low-quality representation \cite{tinyperson, rfla, soda}, while large instances often with clear structures and discriminative features. 
Hence, a series of efforts have been made to boost the semantic representations of small/tiny instances by mining the intrinsic correlations between small and large objects. 
Based on the generative adversarial paradigm, Perceptual GAN \cite{pergan} designs a generator that is optimized to produce high-quality representations of small instances to fool the subsequent discriminator. 
Bai \textit{et al.} \cite{finding} devised a novel pipeline to restore a clear face from the inputting blurry one. 
Noh \textit{et al.} \cite{bffb} further introduced precise supervision for the super-resolution process of small objects. 
Moreover, Wu \textit{et al.} \cite{mimic} and Kim \textit{et al.} \cite{lpr} both exploited similarity learning to force the features of small-scale pedestrians close to that of the large-scale ones which are obtained by an additional model.
The existence of super-resolution branch or offline feature bank hampers the end-to-end optimization while our method updates the exemplar features in an online fashion, which guarantees the diversity of high-quality feature sets thereby getting rid of the collapse issue.

\noindent \textbf{Contrastive Learning for Object Detection.}
The recent explosion of self-supervised learning mainly comes from its Contrastive Learning fashion, and several works have extended this paradigm into detection fields. Detco \cite{detco} is an effective self-supervised framework for object detection which utilizes the image and its local patches to conduct contrastive learning. 
Wu \textit{et al.} \cite{smoky} applied contrastive learning to object detection under smoky conditions. 
Though contrastive learning has recently received considerable interests \cite{simclr, moco, cross}, the potential of utilizing contrastive learning for better representation of small objects has not yet been investigated to date.

\section{Our Method}
This section presents the details about CFINet. 
We start with a discussion about the inherent limitations of Cascade RPN when confronting small objects, then our coarse-to-fine high-quality proposal generation pipeline tailored for size-limited instances is introduced. 
Afterwards, we elucidate the architecture of newly designed Feature Imitation branch, also with the optimization and training procedure.
The overall architecture of CFINet is shown in Figure \ref{fig:Fig2}.

\subsection{Towards Better Proposals}
\noindent \textbf{Limitations of Cascade RPN.}  High-quality proposals play a pivotal role in two-stage detectors, but need heuristic anchor settings. Casdade RPN \cite{cascaderpn} discards this conventional setup by placing one single anchor on each feature point and conducting multi-stage refinement.
Though exhibiting superior performance on objects of general scales, Cascade RPN fails to tackle extremely small objects well due to its inherent limitations. 
Concretely, the distance metric used in first-stage regression cannot guarantee sufficient potential anchors for small objects who have significantly small \textit{center region}. 
Moreover, Cascade RPN only marks eligible anchors on a single pyramid level as \textit{positive}, while this heuristic scheme simply discards those possible anchors at other levels which can still convey the existence and rough location information of small objects \cite{querydet}. 

\noindent \textbf{Coarse-to-fine RPN.}
To remedy the aforementioned issues of Cascade RPN when handling small instances, we propose Coarse-to-fine RPN and the detailed structure is in Figure \ref{fig:Fig2}.
First, we design an area-based anchor selection strategy to enable the instances of various sizes could have (relatively) adequate potential anchors. 
Concretely, for an object box with the width $w$ and the height $h$, any anchors who have an IoU larger than $T_{\mathrm{a}}$ will be regarded as \textit{positive} for the coarse regression, and $T_{\mathrm{a}}$ is formulated as follow:

\begin{equation}\label{eq1}
	T_{\mathrm{a}} = \max(0.25, 0.20+\gamma \cdot \log {\frac{\sqrt{w\cdot h}}{12}}),
\end{equation}

\noindent where $\gamma$ denotes a scale factor and is set to default 0.15 in our experiments, and the term
12 actually corresponds to the minimal area definition of SODA dataset \cite{soda}, which enables adequate samples for extreme-size objects and can be tuned for different datasets.
Moreover, $\gamma$ and $\mathrm{max}$ operation keep the optimization from being overwhelmed by the low-quality priors.
Taking IoU as the criterion to mine potential anchors, the optimization inconsistency in multi-stage regression of Cascade RPN can be averted.
Meanwhile, the model determines the positive sample in a more smooth way on top of the proposed continuous threshold.

Distinct from Cascade RPN, we preserve anchors of all Feature Pyramid Network (FPN) \cite{fpn} levels \{$ P_{\mathrm{2}}, P_{\mathrm{3}}, P_{\mathrm{4}}, P_{\mathrm{5}} $\}  to perform first-stage regression.
In this way, we could mine sufficient potential anchors for extremely small instances and meanwhile, larger instances still can obtain proper attention since the anchors matched to them have naturally higher IoUs, as discussed in Figure \ref{fig:Fig1}. 
After the first-stage regression, we then capture the offsets inside the regressed boxes and input them with the feature maps to RPN, in which the Adaptive Convolution \cite{cascaderpn} will be exploited to align the features and conduct second-stage regression and foreground-background classification.

\noindent \textbf{Loss Function.} 
The training objective of our CRPN is:

\begin{equation}\label{eq2}
	\mathcal{L}_{\mathrm{CRPN}} = {\alpha}_{\mathrm{1}} \left(  \mathcal{L}_{\mathrm{reg}}^{\mathrm{c}} + \mathcal{L}_{\mathrm{reg}}^{\mathrm{f}} \right) + {\alpha}_{\mathrm{2}} \mathcal{L}_{\mathrm{cls}} ,
\end{equation}

\noindent where we use cross-entropy loss and IoU Loss \cite{iouloss} as $\mathcal{L}_{\mathrm{cls}}$ and $\mathcal{L}_{\mathrm{reg}}^{\cdot}$, respectively. The $\mathrm{c}$ and $\mathrm{f}$ in Eq. (\ref{eq2}) indicate the coarse-stage and fine-stage in our CRPN, and noting that we only do classification in the latter stage.
The loss weights ${\alpha}_{\mathrm{1}}$ and ${\alpha}_{\mathrm{2}}$ are set to $9.0$ and $0.9$, respectively.

\subsection{Feature Imitation for Small Object Detection} \label{fi-head}
Efforts on exploiting the intrinsic correlations between objects of different scales to boost the representations of small objects have been made, but most of them fail to the effectiveness and diversity. 
Specifically, the majority of previous methods \cite{pergan, mtgan, finding, bffb} resort to GAN to super-resolve the representations of small instances. 
This calls for the sophisticated training schemes and is prone to fabricate fake textures and artifacts \cite{efpn}. 
Another line of efforts turn to the similarity learning which either has to construct offline feature bank in a cumbersome way \cite{lpr}, or directly leverages $\mathcal{L}_2$ norm to similarity measurement between different RoI features \cite{mimic}, potentially leading the feature collapse issue: the region features after amending could have high-similarity but lost their own characteristics. 
This homogenization in feature space actually impairs the generality and robustness of the model.

To mitigate the collapse risks and avoid the memory burden as well as enable the end-to-end optimization, we devise a Feature Imitation (FI) head (see Figure \ref{fig:Fig2}). 
Most importantly, instead of solely taking large-scale objects as the guidance of this procedure, we consider the model response in current state for each instance, thereby constructing a \textbf{dynamic} and currently \textbf{optimized} feature bank of proper exemplars in an \textbf{online} fashion. 
The FI branch mainly composes an Exemplar Feature Set and a Feature-to-Embedding (Feat2Embed) module, where the former reserves the RoI features of high-quality exemplars and the latter projects the input to the embedding space. 
Next we elucidate the details about our Feature Imitation branch.

\noindent \textbf{What is a proper exemplar?} 
As we discussed above, an exemplar is vital in the imitation learning.
To determine the most representative/proper/high-quality examples which can deliver authentic guidance/supervision for small objects confusing the model at this moment, we first introduce a simple quality indicator for an instance. 
Given a ground-truth (GT) object $g=(c^*, b^*)$, where $c^*$ and $b^*$ denote its label and bounding box coordinates. 
Assuming the detection head outputs a prediction set $\mathcal{S} = \{\bm{\mathit{C}}_i, \mathit{{IoU}_i} \}_{i=1,2,...,M}$ for $g$, in which $\bm{\mathit{C}}_i \in \mathbb{R}^\mathit{N+1}$ indicates the predicted classification vector and ${IoU}_i$ stands for the IoU of predicted box to GT, and $N$ is the number of foreground classes.
Then we can obtain the potential high-quality set $\mathcal{S^{'}} = \{(\bm{\mathit{C}}_j, \mathit{{IoU}_i}) | \arg\max \bm{\mathit{C}}_j=c^* \}_{j=1,2,...,M^{'}}$ where $M^{'} \leq M$. 
Now the Instance Quality of an object $g$ is defined as:

\begin{equation}\label{eq3}
	IQ=\frac{1}{\left| \mathcal{S^{'}} \right|}\sum_{j=1}^{M^{'}}{C_{j,c^*}\cdot}IoU_j
\end{equation}

The $IQ$ of a GT serves as an indicator of the current model's detection capability, enabling us to capture the high-quality exemplars who have precise localization and high-confidence classification scores, and the instances confusing the model often fail to either of them.
By setting appropriate threshold, we can select proper instances to build the teacher feature-set and perform the imitation process.

\begin{algorithm}[t!]
	\caption{ Training of Feature Imitation branch. }
	\label{alg1}
	\begin{algorithmic}[1] 
		\REQUIRE ~~\\
		The set of GT boxes $\mathcal{G} = \{c_i^*, b_i^* \}_{i=1,2,...,T}$ and corresponding RoI features 
		$\{\bm{x}_i^g \}_{i=1,2,...,T}$ in current batch;\\
		The set of exemplar features $\mathcal{E} = \{\mathcal{E}_i \}_{c=1,2,...,N} $; \\
		The set of background RoI features in current batch $\mathcal{X}_{\mathrm{bg}}$; \\
		The threshold of high-quality $T_{\mathrm{hq}}$; \\ 
		The number of pos/neg samples $N_{\mathrm{pos}}$ and $N_{\mathrm{neg}}$; \\
		The transformation function $\mathbf{\Gamma}$; \\
		\ENSURE ~~\\ 
		The set of positive embeddings $\mathcal{P}_{\mathrm{pos}}$ and negative embeddings $\mathcal{P}_{\mathrm{neg}}$ 
		\STATE Initialize the set of positive features $\mathcal{X}_{\mathrm{pos}} $ and negative features $\mathcal{X}_{\mathrm{neg}} $ with $\oslash $;\\
		\FOR {$g$ in $\mathcal{G}$}
		\STATE Compute the $IQ$ of current $g$ according to Eq. (\ref{eq3}) 
		\STATE $\mathcal{X}_{\mathrm{neg}}^g \leftarrow $ sample $N_{\mathrm{neg}}$ features from $ \mathcal{X}_{\mathrm{bg}} \cup \mathcal{E} \backslash \mathcal{E}_{c^*}$
		\IF{$IQ \geq T_{\mathrm{hq}}$}
		\STATE $\mathcal{E}_{c^*} \leftarrow \bm{x}_i^g $ 
		\STATE $\mathcal{X}_{\mathrm{pos}}^g \leftarrow \Gamma(\bm{x}_i^g)$
		\ELSE 
		\STATE $\mathcal{X}_{\mathrm{pos}}^g \leftarrow $ sample $N_{\mathrm{pos}}$ features from $\mathcal{E}_{c^*}$
		\ENDIF
		\STATE $\mathcal{X}_{\mathrm{pos}} = \mathcal{X}_{\mathrm{pos}} \cup \mathcal{X}_{\mathrm{pos}}^g $, $\mathcal{X}_{\mathrm{neg}} = \mathcal{X}_{\mathrm{neg}} \cup \mathcal{X}_{\mathrm{neg}}^g $
		\STATE Apply Eq. (\ref{eq4}) to $\mathcal{X}_{\mathrm{pos}}$ and $\mathcal{X}_{\mathrm{neg}}^g $ to obtain $\mathcal{P}_{\mathrm{pos}}$ and $\mathcal{P}_{\mathrm{neg}}$ 
		\ENDFOR
		\RETURN $\mathcal{P}_{\mathrm{pos}}$ and $\mathcal{P}_{\mathrm{neg}}$
	\end{algorithmic}
\end{algorithm}

\noindent \textbf{Feat2Embed Module.}
Instead of directly measuring the similarity between different RoI features \cite{mimic}, we first embed these features with the simple Feat2Embed module.
The input of FI branch is the region feature $\bm{x}_i \in \mathbb{R}^{H \times W \times C} $ obtained
by the RoI-wise operation, \textit{e.g.}, RoI Align, which will be first processed by three consecutive $3 \times 3$ convolutional layers (with no padding operation) to abstract compact representations. 
It is worth noting that we update parameters during the extraction of current regional features and freeze parameters during the extraction of exemplar ones (see Feat2Embed module in Figure \ref{fig:Fig2}), resulting in improved stability in performance.
Subsequently, the intermediate features will be mapped to the embedding space on top of a two-layer perceptron and the embedding layer with the dimension of 128, in which the dimension of hidden layers is set to 512. 
We have also investigated various design choices and structures for our Feat2Embed module, and detailed information can be found in the Supplementary Materials.
Finally, the output of the Feature Imitation branch is defined as:

\begin{equation}\label{eq4}
	\bm{v}_i = \varTheta_{\mathrm{FI}}(\bm{x}_i),
\end{equation}

\noindent where $\varTheta_{\mathrm{FI}}$ denotes the parameters of Feature Imitation branch to be optimized.

\begin{figure}[t]
	\centering
	\includegraphics[scale=0.475]{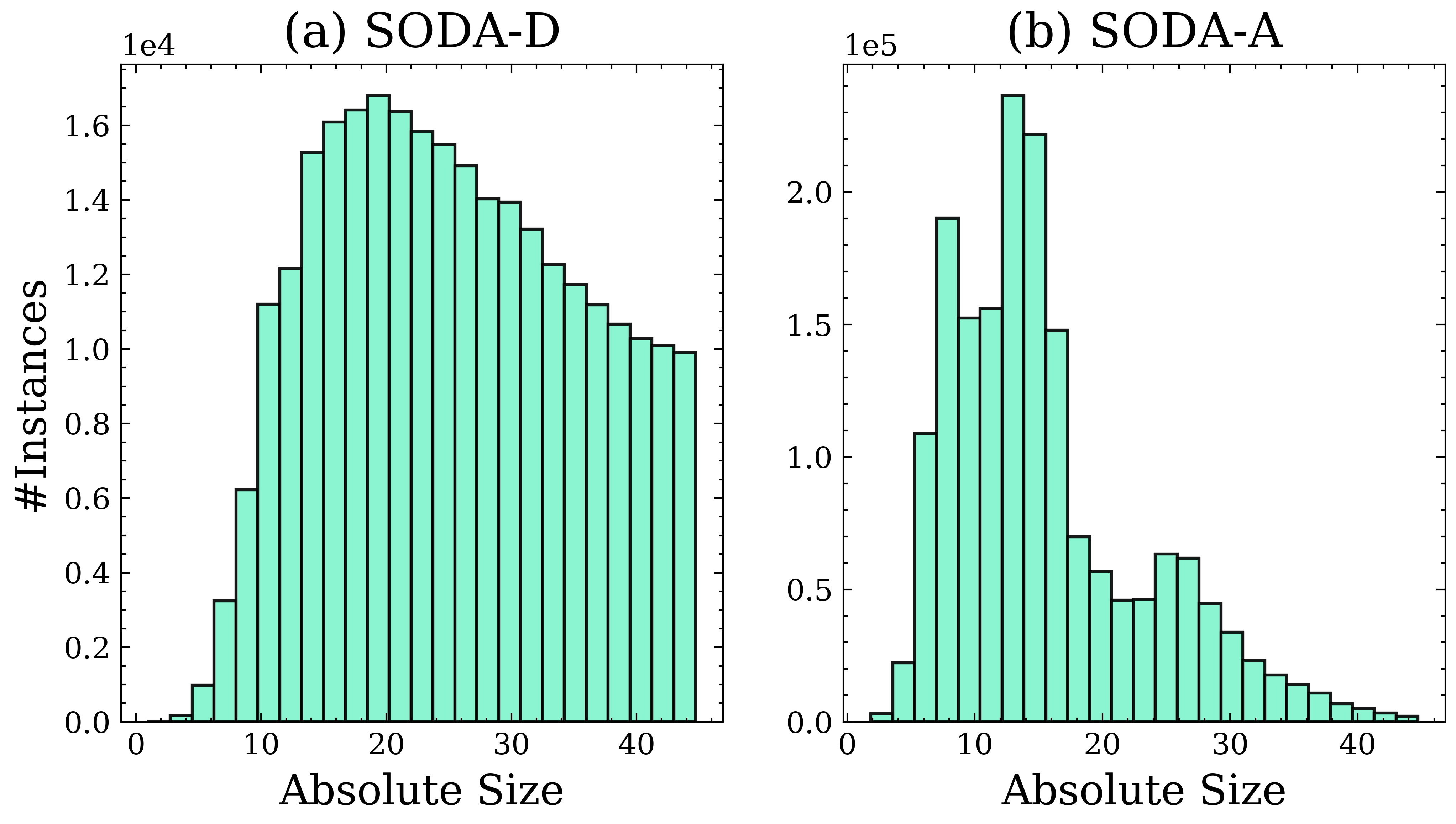}
	\caption{Size distribution of instances in (a) SODA-D and (b) SODA-A, where the absolute size corresponds to the square root of the object area.}
	\label{fig:Fig-avg}
\end{figure}

\begin{table*}[t!]
	\centering
	\resizebox{0.85\textwidth}{!}{
		\begin{tabular}{|c|c|c|ccc|cccc|}
			\hline
			Method & Publication & Schedule & $AP$  & $AP_{50}$  & $AP_{75}$  & $AP_{eS}$  & $AP_{rS}$  & $AP_{gS}$  & $AP_{N}$ \\
			\hline
			\hline
			\textit{\textbf{One-stage}} & \multicolumn{1}{c}{} & \multicolumn{1}{c}{} &       &       & \multicolumn{1}{c}{} &       &       &       &  \\
			\hline
			RetianNet \cite{retinanet} & ICCV'17 & 1$\times$    & 28.2  & 57.6 & 23.7 & 11.9    & 25.2  & 34.1  & 44.2 \\
			FCOS \cite{fcos} & ICCV'19 & 1$\times$    & 23.9  & 49.5  & 19.9    & 6.9  & 19.4  & 30.9  & 40.9  \\
			ATSS \cite{atss} & CVPR'20 & 1$\times$    & 26.8 & 55.6 & 22.1 & 11.7 & 23.9 & 32.2 & 41.3   \\
			YOLOX \cite{yolox} & ArXiv'21 & 70e & 26.7  & 53.4  & 23.0  & 13.6  & 25.1  & 30.9  & 30.4 \\
			DyHead \cite{dyhead} & CVPR'21 & 1$\times$ & 27.5  & 56.1  & 23.2  & 12.4  & 24.4  & 33.0  & 41.9 \\
			\hline
			\textit{\textbf{Keypoint-based}} & \multicolumn{1}{c}{} & \multicolumn{1}{c}{} &       &       & \multicolumn{1}{c}{} &       &       &       &  \\
			\hline
			CornerNet \cite{cornernet} & ECCV'18 & 2$\times$    & 24.6 & 49.5 & 21.7 & 6.5 & 20.5 & 32.2 & 43.8 \\
			CenterNet \cite{points} & ArXiv'19 & 70e    & 21.5  & 48.8  & 15.6  & 5.1  & 16.2  & 29.6  & 42.4 \\
			RepPoints \cite{reppoints} & ICCV'19 & 1$\times$    & 28.0  & 55.6  & 24.7  & 10.1 & 23.8  & 35.1  & \textbf{45.3}  \\
			\hline
			\textit{\textbf{Query-based}} & \multicolumn{1}{c}{} & \multicolumn{1}{c}{} &       &       & \multicolumn{1}{c}{} &       &       &       &  \\
			\hline
			Deformable-DETR \cite{ddetr} & ICLR'20 & 50e   & 19.2  & 44.8  & 13.7  & 6.3  & 15.4  & 24.9  & 34.2  \\
			Sparse RCNN \cite{sparsercnn} & CVPR'21 & 1$\times$ &  24.2 & 50.3 & 20.3 & 8.8 & 20.4 & 30.2 & 39.4  \\
			\hline
			\textit{\textbf{Two-stage}} & \multicolumn{1}{c}{} & \multicolumn{1}{c}{} &       &       & \multicolumn{1}{c}{} &       &       &       &  \\
			\hline
			Baseline \cite{fasterrcnn} & NeurIPS'15 & 1$\times$    & 28.9  & 59.4  & 24.1  & 13.8  & 25.7  & 34.5  & 43.0 \\
			\hdashline[3pt/ 1.5pt]
			Cascade RPN \cite{cascaderpn} & NeurIPS'19 & 1$\times$ & 29.1 & 56.5 & 25.9 & 12.5 & 25.5 & 35.4 & 44.7 \\
			RFLA \cite{rfla} & ECCV'22 & 1$\times$    & 29.7  & 60.2  & 25.2  & 13.2  & 26.9  & 35.4  & 44.6 \\
			\rowcolor{gray!30}
			CFINet (ours) & - & 1$\times$ & \textbf{30.7} & \textbf{60.8} & \textbf{26.7} & \textbf{14.7} & \textbf{27.8} & \textbf{36.4} & 44.6 \\
			\hline
		\end{tabular}
	}
	\vspace{0.5em}
	\caption{Comparison with state-of-the-art detection approaches on the SODA-D \texttt{test-set}, where 'Baseline' refers to Faster RCNN \cite{fasterrcnn}, serving as the baseline for the two-stage methods in the table. All the methods are trained on a ResNet-50 \cite{resnet}, except YOLOX (CSP-Darknet) \cite{yolox} and CornerNet (HourglassNet-104) \cite{cornernet}. 'Schedule' denotes the number of epochs for training, in which '1$\times$' corresponds to 12 epochs and '50e' indicates 50 epochs.}
	\label{tab:Tab2}%
\end{table*}%

\noindent \textbf{Loss Function.}
The objective of our FI head is simple: calculating the similarity between the RoI feature of proposal and that of the stored high-quality instances in embedding space, thereby pulling the features of those instances that confuse the model close to the exemplar ones of belonging category, while pushing apart from that of other categories and backgrounds.
To this end, we propose a loss function based on Supervised Contrastive Learning \cite{scl} which extends the contrastive learning setup and allows multiple positive samples for an anchor object by exploiting the accessible label information. 
The loss function tailored for our FI branch is as follows:

\begin{equation}\label{eq5}
	\mathcal{L} _{\mathrm{FI}}=\frac{-1}{\left| \mathcal{P} _{\mathrm{pos}} \right|}\sum_j{\sum_{v_{\mathrm{p}}\in \mathcal{P} _{\mathrm{pos}}}{\log \frac{\exp \left( v_j\cdot v_{\mathrm{p}}/\tau \right)}{\sum_{v_i\in \mathcal{P}}{\exp \left( v_j\cdot v_i/\tau \right)}}}},
\end{equation}

\noindent where $\mathcal{P} = \mathcal{P}_{\mathrm{pos}} \cup \mathcal{P}_{\mathrm{neg}}$ denotes the sample set, while $\mathcal{P}_{\mathrm{pos}}$ and $\mathcal{P} _{\mathrm{neg}}$ represent the positive and negative set respectively and they have the same cardinality ideally, and $v_{\mathrm{p}}$ and $v_{\mathrm{n}}$ are the positive and negative sample from $\mathcal{P}_{\mathrm{pos}}$ and $ \mathcal{P}_{\mathrm{neg}}$.
Moreover, $j$ indexes the current proposal and $\tau$ indicates the temperature which plays a crucial part in contrastive learning and needs to be well designed, and we conduct ablation studies (see Table \ref{tab:Tab9}) to determine the optimal setting in our framework. 
The total loss function is presented:

\begin{equation}\label{eq6}
	\mathcal{L} = \mathcal{L}_{\mathrm{CRPN}} + \mathcal{L}_{\mathrm{cls}} + \mathcal{L}_{\mathrm{reg}} + {\alpha}_{\mathrm{3}} \mathcal{L}_{\mathrm{FI}},
\end{equation}

\noindent in which $\mathcal{L}_{\mathrm{cls}}$ and $\mathcal{L}_{\mathrm{reg}}$ are the original losses of detection head, and the $\alpha_3$ is utilized to scale the weight of Feature Imitation part.
With contrstive learning setups, not only can we fulfill the imitation learning but also prevent the collapse issue, thereby boosting the representations of small instances effectively.
Moreover, the imitation process is only installed in training phase and will not slow the pace of inference.

\noindent \textbf{Training.}
Next we elucidate the training details of FI branch. 
The exemplar set $\mathcal{E} = \{\mathcal{E}_i \}_{c=1,2,...,N} $ containing high-quality features of $N$ foreground categories and $\mathcal{E}_i = \{\bm{x}_{i,j} \}_{j=1,2,...,N_i} $ corresponding to the exemplar features of the $i$-th class, and $N_i$ represents its size. 
We use $T_{\mathrm{hq}}$ to pick out those high-quality instances which are compatible to be a good exemplar, and in practice, we set a bound value to the number of high-quality predictions of an instance to filter the effect of fluctuation of the network. 
The function $\mathbf{\Gamma}$ is used to augment features for high-quality instances, \textit{i.e.}, the positive features for a high-quality instance are the transformations of itself. 
The overall training procedure of FI branch is shown in Alg. \ref{alg1} and more details please refer to the Supplementary Materials.

\section{Experiments}
\subsection{Dataset}
To evaluate the effectiveness of our method, we perform extensive experiments on the recently released large-scale benchmark tailored for small object detection: SODA \cite{soda}, including SODA-D and SODA-A.

\noindent \textbf{SODA-D}. 
Focusing on the driving scenario, SODA-D comprises 24828 high-quality images and 278433 instances distributed across nine categories: \textit{people}, \textit{rider}, \textit{bicycle}, \textit{motor}, \textit{vehicle}, \textit{traffic-sign}, \textit{traffic-light}, \textit{traffic-camera}, and \textit{warning-cone}. 
One of the most distinctive strengths of SODA-D is its diversity in terms of period, geographical locations, weather conditions, camera viewpoints, \textit{etc.}

\noindent \textbf{SODA-A}. 
SODA-A contains 872069 objects with oriented box annotations in 2513 aerial images and encompassing nine classes: \textit{airplane}, \textit{helicopter}, \textit{small-vehicle}, \textit{large-vehicle}, \textit{ship}, \textit{container}, \textit{storage-tank}, \textit{swimming-pool}, and \textit{windmill}.
The instances in SODA-A can appear in arbitrary orientations and are with significant density variations.
To be specific, the average number of instances per image in SODA-A is about 350.

\begin{table*}[t!]
	\centering
	\resizebox{0.85\textwidth}{!}{
		\begin{tabular}{|c|c|c|ccc|cccc|}
			\hline
			Method & Publication & Schedule & $AP$  & $AP_{50}$  & $AP_{75}$  & $AP_{eS}$  & $AP_{rS}$  & $AP_{gS}$  & $AP_{N}$ \\
			\hline
			\hline
			\textit{\textbf{One-stage}} & \multicolumn{1}{c}{} & \multicolumn{1}{c}{} &       &       & \multicolumn{1}{c}{} &       &       &       &  \\
			\hline
			Rotated RetinaNet \cite{retinanet} & ICCV'17 & 1$\times$    & 26.8  & 63.4  & 16.2  & 9.1  & 22.0  & 35.4  & 28.2  \\
			S$^2$A-Net \cite{s2anet} & TGRS'22 & 1$\times$    & 28.3  & 69.6  & 13.1  & 10.2  & 22.8  & 35.8 & 29.5  \\
			Oriented RepPoints \cite{orientedrep} & CVPR'22 & 1$\times$  & 26.3  & 58.8  & 19.0  & 9.4  & 22.6  & 32.4  & 28.5  \\
			DHRec \cite{dhrec} & TPAMI'22 & 1$\times$  & 30.1 & 68.8 & 19.8 & 10.6 & 24.6 & 40.3 & 34.6  \\
			\hline
			\textit{\textbf{Two-stage}} & \multicolumn{1}{c}{} & \multicolumn{1}{c}{} &       &       & \multicolumn{1}{c}{} &       &       &       &  \\
			\hline
			Baseline \cite{fasterrcnn} & NeurIPS'15 & 1$\times$    & 32.5 & 70.1  & 24.3  & 11.9  & 27.3  & 42.2 & 34.4  \\
			\hdashline[3pt/ 1.5pt]
			Gliding Vertex \cite{gliding} & TPAMI'21 & 1$\times$    & 31.7 & 70.8 & 22.6 & 11.7 & 27.0 & 41.1 & 33.8  \\
			Oriented RCNN \cite{orientedrcnn} & ICCV'21 & 1$\times$    & \textbf{34.4}  & 70.7  & \textbf{28.6}  & 12.5  & 28.6  & \textbf{44.5} & \textbf{36.7}  \\
			DODet \cite{dodet} & TGRS'22 & 1$\times$  & 31.6 & 68.1 & 23.4 & 11.3 & 26.3 & 41.0 & 33.5  \\
			\rowcolor{gray!30}
			CFINet (ours) & - & 1$\times$  & \textbf{34.4} & \textbf{73.1} & 26.1 & \textbf{13.5} & \textbf{29.3} & 44.0 & 35.9 \\
			\hline
		\end{tabular}
	}
	\vspace{0.5em}
	\caption{Comparison with state-of-the-art detection approaches on the SODA-A \texttt{test-set}, where 'Baseline' refers to Rotated Faster RCNN \cite{fasterrcnn}, serving as the baseline for the two-stage methods in the table. Other settings are consistent with Table \ref{tab:Tab2}.
	}
	\label{tab:Tab3}%
	\vspace{0.25em}
\end{table*}%

\begin{figure*}[t]
	\centering
	\includegraphics[scale=0.6]{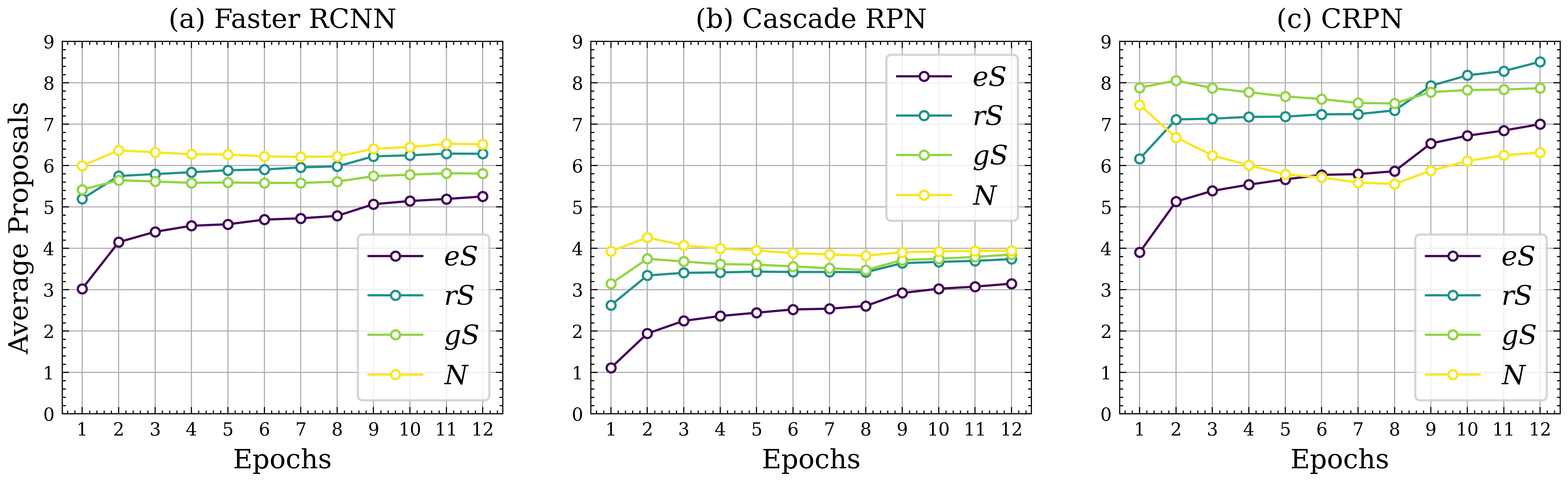}
	\caption{The average number of high-quality proposals generated by (a) RPN, (b) Cascade RPN and (c) CRPN for instances in \textit{extremely Small} (\textit{eS}), \textit{relatively Small} (\textit{rS}), \textit{generally Small} (\textit{gS}), and \textit{Normal} (\textit{N}) subsets, respectively. Noting that a proposal who has an IoU larger than $0.5$ to any ground-truth boxes will be registered as a high-quality proposal.}
	\label{fig:Fig3}
	\vspace{0.25em}
\end{figure*}

\begin{table}[t]
	\centering
	\small
		\begin{tabular}{|c|c|cccc|}
			\hline
			Proposal Method & $AR$  & $AR_{eS}$  & $AR_{rS}$  & $AR_{gS}$  & $AR_N$  \\
			\hline
			\hline
			RPN \cite{fasterrcnn}  & 41.2 & 24.0 & 38.3 & 47.3 & \textbf{57.1}   \\
			RPN-0.5  & 41.3 & 24.2 & 38.5 & 47.3 & 54.1   \\
			GA-RPN \cite{garpn} & 42.1 & 24.1 & \textbf{39.2} & 48.9 & 56.2 \\
			Cascade RPN \cite{cascaderpn} & 41.8 & 22.8 & 38.2 & 48.7 & \textbf{57.1}  \\
			\rowcolor{gray!30}
			CRPN & \textbf{42.6} & \textbf{24.6} & 38.9 & \textbf{49.1} & 56.9 \\
			\hline
		\end{tabular}
	\vspace{0.25em}
	\caption{Average Recall ($AR$) performances of our CRPN and its counterparts on the SODA-D \texttt{test-set}. All the methods are with Faster RCNN (ResNet-50) as the baseline and trained for a $1\times$ schedule, in which RPN denotes the vanilla version of Faster RCNN whose positive threshold in RPN stage is set $0.7$, and RPN-0.5 represents the version with 0.5 as its positive threshold of RPN. The results are tested with 300 proposals per image.}
	\label{tab:Tab4}%
\end{table}%

As a specialized benchmark for small object detection, the instances in SODA are tiny, with most of them having an average size ranging from 10 to 30 pixels (see Figure \ref{fig:Fig-avg}). 
In contrast to conventional datasets for object detection, SODA includes extensive \textit{ignore} annotations, aimed at filtering out instances that are either too large or are challenging to be identified deterministically due to heavy occlusion or lens flare.
This procedure helps the model focus on valuable small instances.
The objects in SODA are divided into \textit{Small} and \textit{Normal} according to their areas, in which \textit{Small} is further split into three subsets: \textit{extremely Small} (\textit{eS}), \textit{relatively Small} (\textit{rS}) and \textit{generally Small} (\textit{gS}). 
The evaluation metric of SODA follows that of COCO \cite{coco}, namely averaging the precision over 10 IoU thresholds ranging from 0.5 to 0.95 (with an interval of 0.05), specifically focusing on \textit{Small} objects.

\begin{table}[t]
	\centering
	\small
		\begin{tabular}{|ccc|cccc|}
			\hline
			Baseline & CRPN  & FI  & $AP$  & $AP_{eS}$  & $AP_{rS}$  & $AP_{gS}$  \\
			\hline
			\hline
			\checkmark  &  &  & 28.9 & 13.8  & 25.7  & 34.5  \\
			\checkmark  & \checkmark &   & 30.3	 & 14.3  & 27.3 & 36.1 \\
			\checkmark  &  & \checkmark & 29.5 & 14.4 & 26.3 & 35.1 \\
			\checkmark  & \checkmark & \checkmark & \textbf{30.7} & \textbf{14.7} & \textbf{27.8} & \textbf{36.4} \\
			\hline
		\end{tabular}
	\vspace{0.5em}
	\caption{Ablation analysis of our method, in which 'Baseline' denotes the vanilla Faster RCNN, CRPN and FI indicate Coarse-to-fine RPN and Feature Imitation branch, respectively.}
	\label{tab:Tab5}%
	\vspace{-1em}
\end{table}%

\subsection{Implementation Details}
In the following experiments, unless specified, we use \texttt{train-set} to conduct the training and leave \texttt{test-set} to performance comparisons and ablation studies. 
Considering that the images in SODA enjoy a very high resolution ($\sim 4000 \times 3000$), we first split the original images into a series of $800 \times 800$ patches with a stride of 650, and similar to \cite{soda} these patches will be resized to $1200 \times 1200$ during training and testing. 
All the experiments in this paper are conducted on a single RTX 3090 with the batch size of 4. 
Only random flip involved in data augmentation. We train all the models with a $1 \times$ schedule (a bunch of 12 epochs), and the learning rate is set to 0.01 which decays after epoch 8 and epoch 11 by 0.1. The default optimizer is SGD with the momentum of 0.9 and the weight decay of 0.0001. 
We use ResNet-50 \cite{resnet} with FPN \cite{fpn} for all models.

\subsection{Main Results}
To exhibit the effectiveness of our method, we conduct a thorough comparison with current representative approaches on the SODA-D and SODA-A. 

Table \ref{tab:Tab2} represents the results of our method and several mainstream approaches on the SODA-D \texttt{test-set}. 
Integrating with Faster RCNN \cite{fasterrcnn}, our CFINet achieves state-of-the-art performance with an overall $AP$ of $30.7 \%$, and outperforms the baseline model with $1.8 \%$ points. 
When delving into specific metrics, our method exhibits clear predominance, particularly on the most challenging metrics $AP_{eS}$ and $AP_{rS}$.
Moreover, CFINet exceeds the tailored small object detection method RFLA \cite{rfla} by a significant margin ($1.0\%$ on $AP$, $1.5\%$ on $AP_{es}$, and $0.9\%$ on $AP_{rS}$) when both taking Faster RCNN as the baseline. 
Actually, RFLA sacrifices the performance on \textit{extremely Small} instances (with a decrease of $0.6 \%$ points when compared to Faster RCNN).

On the SODA-A \texttt{test-set}, CFINet also achieves the best result and shows great advantage in comparison to other solutions especially on $AP_{eS}$ (see Table \ref{tab:Tab3}), indicating its superiority and generality.
Furthermore, albeit  exhibiting advantage on $AP_{75}$ metric and larger instances, Oriented RCNN \cite{orientedrcnn} lags largely behind our approach on $AP_{50}$ ($73.1 \%$ \textit{vs.} $70.7 \%$) and $AP_{eS}$ ($13.5 \%$ \textit{vs.} $12.5 \%$).

\begin{table}[t]
	\centering
	\small
		\resizebox{0.7\columnwidth}{!}{
		\begin{tabular}{|c|cccc|}
			\hline
			Strategy  & $AP$  & $AP_{eS}$  & $AP_{rS}$  & $AP_{gS}$  \\
			\hline
			\hline
			0.20  & 29.9 & 13.7 & 26.8 & 35.9 \\
			0.40 & 30.1 & 14.1 & 26.9 & \textbf{36.2} \\
			Ours & \textbf{30.3} & \textbf{14.3}  & \textbf{27.3} & 36.1 \\
			\hline
		\end{tabular}%
		}
	\vspace{0.5em}
	\caption{Different definitions about \textit{positive} anchor for CRPN, in which 'Ours' denotes the proposed dynamic strategy.}
	\label{tab:Tab6}%
\end{table}%

\begin{table}[t]
	\centering
	\small
		\resizebox{0.7\columnwidth}{!}{
		\begin{tabular}{|c|cccc|}
			\hline
			$\alpha_{\mathrm{3}}$  & $AP$  & $AP_{eS}$  & $AP_{rS}$  & $AP_{gS}$   \\
			\hline
			\hline
			0.25 & 30.6 & \textbf{14.7} & 27.6 & 36.2 \\
			0.50 & \textbf{30.7} & \textbf{14.7}  & \textbf{27.8} & \textbf{36.4} \\
			0.75 &  30.4 & 14.0 & 27.5 & 36.2 \\
			\hline
		\end{tabular}%
			}
	\vspace{0.5em}
	\caption{The effect of loss weight for Feature Imitation branch.}
	\label{tab:Tab7}%
	\vspace{-1em}
\end{table}%

\subsection{Effectiveness of CRPN}
One of the main designs in this paper is the Coarse-to-fine RPN, which is based on the observation that current fixed overlaps-based sampling paradigm is inappropriate for small instances due to the inherent contradictions, while the refined designs of RPN could partially reduce this barrier but still fail to satisfactory results. 
Here, we conduct thorough analyses to demonstrate the capability of our CRPN to generate high-quality proposals for size-limited instances.

We first exhibit the recall performances of our CRPN and its counterparts in Table \ref{tab:Tab4}, from which we can see that lowering the positive threshold slightly improves the average recall while sacrificing the performance of larger instances ($AR_N$ experiences a sharp decline from $57.1\%$ to $54.1\%$). GA-RPN \cite{garpn} as well as Cascade RPN \cite{cascaderpn} both fail to better results since their patterns incline to large instances as we discussed before. 
In comparison with RPN and its variants, our CRPN demonstrates superior performance on objects in the \textit{Small}, while achieving comparable results on \textit{Normal} instances. 
This validates our assumption that refined proposal networks tend to favor larger objects.

We conjecture that one of the most challenging issues towards accurate small object detection is the scarcity of high-quality samples, which is also the major motivation behind the design of CRPN. 
Hence, we intuitively compare the baseline RPN, Cascade RPN and our CRPN about the number of high-quality samples. 
In Figure \ref{fig:Fig3}, our CRPN generates more high-quality proposals compared to the other competitors.
Interestingly, \textbf{CRPN can dynamically shift the focus along the training}: at the beginning, the model concentrates more on large objects which are conducive for early optimization while as the training goes, the model gradually shifts its attention to objects with small sizes that are usually not handled well before. 
This is interpretable since the instances having extremely limited sizes are with more uncertainties and fitting them in early-phase is not an optimal choice for the detector.

\begin{table}[t]
	\centering
	\small
	\resizebox{0.65\columnwidth}{!}{
		\begin{tabular}{|c|cccc|}
			\hline
			$T_{\mathrm{hq}}$  & $AP$  & $AP_{eS}$  & $AP_{rS}$  & $AP_{gS}$   \\
			\hline
			\hline
			0.50 & 30.3 & 14.0 & 27.3 & 36.0 \\
			0.55 & 30.6 & 14.3 & 27.3 & \textbf{36.5} \\
			0.65 & \textbf{30.7} & \textbf{14.7}  & \textbf{27.8} & 36.4 \\
			0.70 & 30.5 & 14.6 & 27.3 & 36.3 \\
			\hline
		\end{tabular}%
	}
	\vspace{0.5em}
	\caption{The investigation of the criterion to be an exemplar instance.}
	\label{tab:Tab8}%
\end{table}%

\begin{table}[t]
	\centering
	\small
	\resizebox{0.65\columnwidth}{!}{
		\begin{tabular}{|c|cccc|}
			\hline
			$\tau$  & $AP$  & $AP_{eS}$  & $AP_{rS}$  & $AP_{gS}$  \\
			\hline
			\hline
			0.10 & 30.3 & 14.1 & 27.3 & 36.1 \\
			0.50 & 30.4 & 14.4 & 27.2 & 36.2 \\
			0.60 & \textbf{30.7} & \textbf{14.7} & \textbf{27.8} & \textbf{36.4} \\
			0.80 & 30.2 & 14.0 & 27.3 & 36.0 \\
			\hline
		\end{tabular}%
	}
	\vspace{0.5em}
	\caption{The choices of temperature $\tau$ to the final performance.}
	\label{tab:Tab9}%
\end{table}%

\begin{figure*}[t!]
	\centering
	\includegraphics[scale=1]{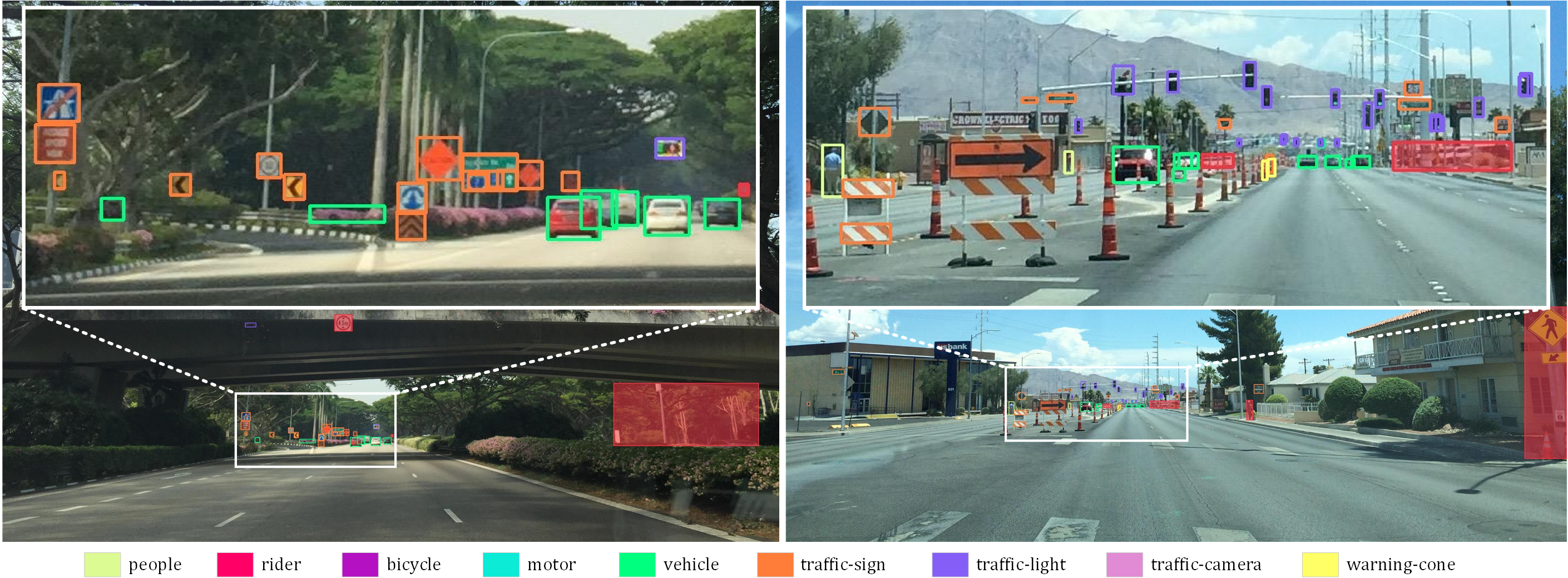} 
	\caption{Qualitative results of our method on the SODA-D \texttt{test-set}. Only predictions with confidence scores larger than 0.3 are demonstrated and the masked bounding boxes represent \textit{ignore} regions. Best viewed in color and zoom-in windows.}
	\label{fig:Fig4}
\end{figure*}

\subsection{Ablation and Discussion}
In this part, we conduct ablation studies as well as comprehensive discussions to attest the importance of the CRPN and FI branch, and moreover, determine the appropriate settings of our approach. All the experiments of this section are conducted on the SODA-D \texttt{test-set}.

\noindent \textbf{Investigation of Designed Components.}
We first perform ablation experiments to verify the effectiveness of two modules. 
As in Table \ref{tab:Tab5}, our CRPN and FI can both improve the performances steadily, while the introduction of feature imitation branch is conducive for the recognition of size-limited instances. 
Consistently, the integration of FI and CRPN achieves the best result, as CRPN is capable of  generating sufficient high-quality proposals (as shown in Figure \ref{fig:Fig3}) thereby offering a more accurate indication of instance quality and the potential to act as an exemplar.

\noindent \textbf{Fixed or Dynamic.}
A natural idea is directly setting fixed \textit{positive} threshold to obtain more anchors for one-stage regression of CRPN. 
Here we show that our simple-yet-effective area-based anchor mining strategy can achieve the best performance. 
In Table \ref{tab:Tab6}, when the IoU threshold of a \textit{positive} potential anchor for first-stage regression drops to $0.20$, the $AP_{eS}$ is only $13.7 \%$ and this could be attributed to low-quality samples and the predominance of large instances as discussed before. 
The proposed area-based anchor mining strategy could mitigate this problem and obtain the best overall accuracy.

\noindent \textbf{Weights of Feature Imitation Loss.} 
In this section, we analyze the impact of the weight parameter of FI branch (namely the hyper-parameter $\alpha_{\mathrm{3}}$) on the model. 
As shown in Table \ref{tab:Tab7}, paying too little or excessive attention both deteriorates the final performance, hence we set $\alpha_{\mathrm{3}}$ to $0.5$ in our experiments to ensure overall accuracy.

\noindent \textbf{The Criterion of Being An Exemplar.}
The quality of the exemplar plays a pivotal role in the imitation process \cite{lpr, mimic}. 
Next we discuss the choices for capturing a high-quality exemplar to build the feature set.
In Table \ref{tab:Tab8}, the lower $T_{\mathrm{hq}}$ involves more exemplars and updates the teacher set more frequently while the higher $T_{\mathrm{hq}}$ does exactly the opposite. 
It can be seen that increasing the $T_{\mathrm{hq}}$ from $0.5$ to $0.65$ could facilitate the imitation process and when the $T_{\mathrm{hq}}$ reaches $0.70$, the overall $ AP$ drops instead. 
This may originates that the earliest samples stored in the feature set are inappropriate for current state, because the optimization is dynamic and the model is evolving hence the criterion of being an exemplar has already changed.

\noindent \textbf{The Temperature.} 
The temperature is paramount for contrastive learning \cite{scl} and we conduct a series of experiments to verify the best choice for $\tau$. From Table \ref{tab:Tab9}, when $\tau$ ranges from $0.10$ to $0.80$, the overall performance increases first and then decreases to $30.2 \%$, therefore we choose $0.6$ for our method.

\noindent \textbf{Visualization.} We demonstrate the visualization results of example images from SODA-D \texttt{test-set} in Figure \ref{fig:Fig4} to intuitively show the capability of our detector when detecting the small instances.

\section{Conclusion}
In this paper, we proposed CFINet, a two-stage detector based on the Coarse-to-fine Region Proposal Network and Feature Imitation setups, in which the former can produce sufficient high-quality proposals for small instances particularly for those with extremely limited sizes. 
Then the novel detection head on top of the feature imitation branch facilitates the representations of small objects posing challenges to the model under the contrastive learning paradigm.
The experiments results show that our method achieves state-of-the-art performance on the large-scale small object detection datasets SODA-D and SODA-A.
In the future, a more flexible and general indicator of instance quality is worth investigating.

\section*{Acknowledgments}
This work was supported in part by the National Science Foundation of China under Grant 62136007 and Grant U20B2068, and in part by the Natural Science Basic Research Program of Shaanxi under Grants 2021JC-16 and 2023-JC-ZD-36.

{\small
	\bibliographystyle{ieee_fullname}
	\bibliography{0817}
}

\newpage

\begin{appendices}
	\setcounter{table}{0}
	\setcounter{figure}{0}
	\renewcommand{\thetable}{\Alph{section}\arabic{table}}
	\renewcommand\thefigure{\Alph{section}\arabic{figure}}
	
	\section{Overview}
	This supplementary material is intended to improve the clarity and comprehensibility of our research. It primarily provides in-depth information about the training procedure and the construction of the exemplar set within the Feature Imitation branch.
	Finally, we describe the empirical limitations about our approach.
	
	\section{Details of Feature Imitation Branch}
	This part we elucidate the detailed settings of training the proposed Feature Imitation (FI) branch, including the policy of producing augmentations for high-quality samples and further discussions about non-high-quality samples, as well as the details about constructing and updating the exemplar feature set.
	
	\noindent \textbf{The Augmentation for High-quality Instances.} 
	In the \textbf{Training} part of Sec. \ref{fi-head} of our main paper, we refer to that the imitation process for a high-quality instance is performed between the feature of itself and its transformed features. 
	In self-supervised contrastive learning paradigm, the only single \textit{positive} sample for an image is generated by the transformation (\textit{e.g.}, AutoAugment \cite{autoaugment}, RandAugment \cite{randaugment} and SimAugment \cite{simclr}).
	Inspired by this setting, a function $\mathbf{\Gamma}$ is employed in our FI head to augment the features for high-quality instances who have an $IQ \geq T_{\mathrm{hq}}$.
	Specifically, we use random translation and zoom-in/out operation to augment the target features, and the corresponding functions are defined as $\mathbf{R}(s_\mathrm{w}, s_\mathrm{h})$ and $\mathbf{Z}(s_{\mathrm{min}}, s_{\mathrm{max}})$, where $s_\mathrm{w}$ and $s_\mathrm{h}$ represent the translation factors along the width-axis and height-axis of the ground-truth box respectively, while $s_{\mathrm{min}}$ and $s_{\mathrm{max}}$ indicate the minimum and maximum factors during zoom-in/out operation. Finally, the overall transformation function $\mathbf{\Gamma}$ is formulated as:
	
	\begin{equation}\label{eq1}
		\mathbf{\Gamma}(x, y, w, h) = \left\{ \mathbf{R} (x, y, w, h), \mathbf{Z} (x, y, w, h) \right\},
	\end{equation}
	
	\noindent where $(x, y, w, h)$ determines the region of proposal.
	In our practices, we use 8 pairs of $(s_\mathrm{w}, s_\mathrm{h})$ and 8 pairs of $(s_{\mathrm{min}}, s_{\mathrm{max}})$ to obtain 16 \textit{positive} samples for a high-quality instance.
	The other transformations may bring better performance while we leave the future work to explore the optimal transformation functions, since the designed simple $\mathbf{\Gamma}$ could fulfill the imitation learning procedure for high-quality instances.
	
	\begin{table}[t]
		\centering
		\small
			\begin{tabular}{|c|cccc|}
				\hline
				$\beta_{\mathrm{1}}, \beta_{\mathrm{2}}, \beta_{\mathrm{3}} $  & $AP$  & $AP_{eS}$  & $AP_{rS}$  & $AP_{gS}$  \\
				\hline
				\hline
				Baseline  & 28.9 & 13.8  & 25.7  & 34.5 \\
				0.5, 0, 0 & 29.0 & 13.7 & 25.7 & 34.7  \\
				0.5, 0.1, 0.05 & \textbf{29.5} & \textbf{14.4} & \textbf{26.3} & \textbf{35.1}  \\
				0.5, 0.2, 0.1 & 29.2 & 14.3 & 26.1 & 34.8 \\
				\hline
			\end{tabular}%
		\vspace{0.5em}
		\caption{The effect of different weights of non-high-quality instances to the performance, where $\beta_{\mathrm{1}}$, $\beta_{\mathrm{2}}$ and $\beta_{\mathrm{3}}$ represent the loss weights of low-quality, mid-quality, and high-quality instances, respectively. 'Baseline' denotes Faster RCNN \cite{fasterrcnn}.}
		\label{tab:Tabs1}%
	\end{table}%
	
	\begin{figure*}[t]
		\centering
		\includegraphics[scale=1]{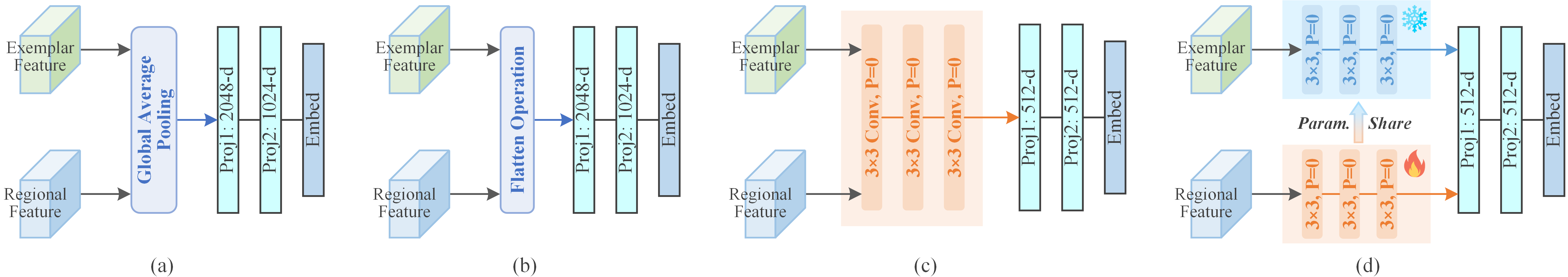}
		\caption{Four architectures for Feat2Embed module: (a) GAP-Embed, (b) Flatten-Embed, (c) Conv-Embed, and (d) SharedConv-Embed.}
		\label{fig:Figs1}
	\end{figure*}
	
	\noindent \textbf{Discussions about Non-high-quality Instances.} 
	We use $IQ$ to indicate the quality of an instance and its competence to be an exemplar by setting a threshold $T_{\mathrm{hq}}$ (practically set to $0.65$), which implies that instances whose $IQ$ scores are below the predefined $T_{\mathrm{hq}}$ are marked as \textbf{low-quality}. Is this reasonable?
	Two instances with the scores of $0.64$ and $0.14$ will be regarded equally as low-quality samples and conduct the imitation, however our core idea of introducing $IQ$ is to mine the exemplars to guide the representation learning of samples with uncertain predictions.
	In other words, these two instances both will be marked as \textit{uncertain}/\textit{ambiguous}, and this is not rigorous because the prediction (classification scores and localization) of the former one ($IQ=0.64$) is actually not bad.
	Hence, to mitigate this issue, we experimentally involve a low-quality threshold $T_{\mathrm{lq}}$ to discover those instances with high demand to be amended.
	Noting the introduction of $T_{\mathrm{lq}}$ will not change the overall training procedure depicted in Alg. 1 of our main paper, and the only difference lies in that we highlight the feature leaning of low-quality instances by assigning different loss weight to instances with a quality score $T_{\mathrm{lq}} \leq IQ < T_{\mathrm{hq}}$ (noted as mid-quality instances) and that with $IQ < T_{\mathrm{lq}}$ (noted as low-quality instances).
	Specifically, we conduct a series experiments to investigate the effect of such settings to the overall performance.
	As in Table \ref{tab:Tabs1}, it is interesting that only focusing on the low-quality instances does not get the best results, and we conjecture this originates that the Feat2Embed module has not been optimized well with low-quality instances only, especially at early stage. 
	Meanwhile, the undue concentration on those non-low-quality instances also poses negative impact to the learning of Feature Imitation branch.
	To sum up, the introduction of mid-quality instances can be regarded as a buffer area that is beneficial for stabilizing the training process and amending the representations of low-quality instances.
	
	\noindent \textbf{Details about the Exemplar Feature Set.}
	The exemplar feature set is crucial in our method, and here we describe some details about its construction and updating rules.
	We empirically set the number of the samples for each ground-truth instance as 128, with half positive samples and half negative ones (except for high-quality instances).
	Moreover, the general rule of updating the exemplar set is resemble that of queue, namely \textit{first in first out}. 
	And the maximum size of the feature set for each category is 256 which is double to that of the sampling number for each instance.
	For the classes with limited high-quality ground truths, we halve the size of exemplar feature set and  positive number to avoid that the feature set is unable to update for a long time.
	
	\noindent \textbf{Choices for Feat2Embed Module.}
	In the Feature Imitation branch, we propose to measure the similarity between different RoI features in the embedding space with the help of the Feat2Embed module. Here, we explore the impact of different Feat2Embed designs on the performance of the FI branch. As demonstrated in Figure \ref{fig:Figs1}, we investigate four pipelines to perform the embedding process: (a) GAP-Embed, (b) Flatten-Embed, (c) Conv-Embed, and (d) SharedConv-Embed. 
	These four architectures consist of two key components: dimensionality reduction and the embedding function. 
	The primary difference among them lies in how they map the regional features to compact representations within the embedding space. 
	We then utilize Faster RCNN as the baseline detector and conduct experiments to identify the optimal setting for the Feat2Embed module. 
	
	\begin{table}[t]
		\centering
		\small
			\begin{tabular}{|c|cccc|}
				\hline
				Feat2Embed  & $AP$  & $AP_{eS}$  & $AP_{rS}$  & $AP_{gS}$  \\
				\hline
				\hline
				Baseline  & 28.9 & 13.8  & 25.7  & 34.5 \\
				GAP-Embed & 29.2 & 14.1 & 25.8 & 34.9  \\
				Flatten-Embed  & 29.4  & \textbf{14.4} & 26.1 & \textbf{35.2} \\
				Conv-Embed  & \textbf{29.5}  & 14.2 & \textbf{26.3} & \textbf{35.2} \\
				SharedConv-Embed & \textbf{29.5} & \textbf{14.4} & \textbf{26.3} & 35.1  \\
				\hline
			\end{tabular}%
		\vspace{0.5em}
		\caption{The effect of different Feat2Embed module designs to the performance of Feature Imitation branch, in which the term 'Baseline' denotes Faster RCNN \cite{fasterrcnn}.}
		\label{tab:Tabs2}%
	\end{table}%
	
	Table \ref{tab:Tabs2} reveals that the proposed Feature Imitation branch demonstrates robustness to most designs except for GAP-Embed. 
	We suspect that directly pooling the regional feature into a single vector results in significant information loss, thereby compromising the representation and similarity computation in the embedding space.
	Given that the number of parameters to be optimized in Flatten-Embed is approximately 60 times that of SharedConv-Embed, and the latter achieves a better average precision ($AP_{eS}$) performance compared to Conv-Embed, we choose SharedConv-Embed as our standard Feat2Embed module.
	
	\noindent \textbf{Empirical Limitations}.
	Albeit facilitating the result of baseline detector on small objects especially on size-limited ones, the Feature Imitation branch may exhibit instability in performance.
	Empirically, the final performance of our feature imitation head significantly relies on the exemplars which dominate the imitation learning.
	However, the exemplar feature set constructed in each training procedure is distinct due to the dynamic of optimization.
	In other words, the exemplar features in current turn may fail to reach the bar of a high-quality teacher feature in next turn, and vice versa.
	Hence, a more flexible and general indicator of instance quality greatly contributes to a more elegant and effective method, and we leave this issue open to further research.
\end{appendices}

\end{document}